\documentclass[sigconf]{acmart}
\usepackage{graphicx} 
\usepackage{subcaption}

\usepackage{makecell}
\usepackage{amsfonts}
\usepackage{amsmath} 
\usepackage[linesnumbered,ruled]{algorithm2e}
\let\oldnl\nl
\newcommand{\nonl}{\renewcommand{\nl}{\let\nl\oldnl}}

\usepackage{bm}

\def\BibTeX{{\rm B\kern-.05em{\sc i\kern-.025em b}\kern-.08emT\kern-.1667em\lower.7ex\hbox{E}\kern-.125emX}}
    
\copyrightyear{2019}
\acmYear{2019}
\setcopyright{acmlicensed}
\acmConference[Alaska '19]{Alaska '19: ACM International Conference on Knowledge Discovery and Data Mining}{August 04--08, 2019}{Alaska, US}
\acmBooktitle{Alaska '19: ACM International Conference on Knowledge Discovery and Data Mining, August 04--08, 2019, Alaska, US}

\begin{document}

%
\title{Deep Uncertainty Quantification: A Machine Learning Approach for Weather Forecasting}

%
\author{Bin Wang}
\affiliation{%
\institution{Centre for Artificial Intelligence,}
  \institution{University of Technology Sydney}
    \institution{Southwest Jiaotong University}
}
\email{bin.wang-7@student.uts.edu.au}

\author{Jie Lu}
\affiliation{%
\institution{Centre for Artificial Intelligence,}
  \institution{University of Technology Sydney}
}
\email{jie.lu@uts.edu.au}

\author{Zheng Yan}
\affiliation{%
\institution{Centre for Artificial Intelligence,}
  \institution{University of Technology Sydney}
  }
\email{yan.zheng@uts.edu.au}

\author{Huaishao Luo}
\affiliation{%
  \institution{Southwest Jiaotong University}
  }
\email{infocom525@gmail.com}

\author{Tianrui Li}
\affiliation{%
\institution{Institute of Artificial Intelligence,}
  \institution{Southwest Jiaotong University}
  \city{Chengdu 611756}
  \country{China}
  }
\email{trli@swjtu.edu.cn}

\author{Yu Zheng}
\affiliation{%
  \institution{Urban Computing Business Unit,}
  \institution{JD Digits}
  \city{Beijing}
  \country{China}
  }
\email{msyuzheng@outlook.com}

\author{Guangquan Zhang}
\affiliation{%
\institution{Centre for Artificial Intelligence,}
 \institution{University of Technology Sydney}
 }
\email{guangquan.zhang@uts.edu.au}
%

%
\begin{abstract}
Weather forecasting is usually solved through numerical weather prediction (NWP), which can sometimes lead to unsatisfactory performance due to inappropriate setting of the initial states. In this paper, we design a data-driven method augmented by an effective information fusion mechanism to learn from historical data that incorporates prior knowledge from NWP. We cast the weather forecasting problem as an end-to-end deep learning problem and solve it by proposing a novel negative log-likelihood error (NLE) loss function.  A notable advantage of our proposed method is that it simultaneously implements single-value forecasting and uncertainty quantification, which we refer to as \textit{deep uncertainty quantification} (DUQ). Efficient deep ensemble strategies are also explored to further improve performance. This new approach was evaluated on a public dataset collected from weather stations in Beijing, China. Experimental results demonstrate that the proposed NLE loss significantly improves generalization compared to mean squared error (MSE) loss and mean absolute error (MAE) loss. Compared with NWP, this approach significantly improves accuracy by 47.76\%, which is a state-of-the-art result on this benchmark dataset. The preliminary version of the proposed method won 2nd place in an online competition for daily weather forecasting \footnote[1]{\textit{AI Challenger 2018}  \url{https://challenger.ai/competition/wf2018}}. 
\end{abstract}

%
%


%

%
\maketitle

\section{Introduction}
Meteorological elements, such as temperature, wind and humidity, profoundly affect many aspects of human livelihood \cite{gneiting2005weather, murphy1993good}. They provide analytical support for issues related to urban computing such as traffic flow prediction, air quality analysis, electric power generation planning and so on \cite{zheng2014urban}. The most common method currently utilized in meteorology is the use of physical models to simulate and predict meteorological dynamics known as numerical weather prediction, or NWP. The advantage of NWP is that it is based on the numerical solution of atmospheric hydro thermo dynamic equations and is able to obtain high prediction accuracy if the initial solution is appropriately chosen. However, NWP may not be reliable due to the instability of these differential equations \cite{tolstykh2005some}. With the growing availability of meteorological big data, researchers have realized that introducing data-driven approaches into meteorology can achieve considerable success. Several machine learning methods have been applied to weather forecasting \cite{sharma2011predicting, grover2015deep, hernandez2016rainfall}. The merit of data-driven methods is that they can quickly model patterns through learning to avoid solving complex differential equations. Nevertheless, learning from historical observations alone requires big data and a tedious amount of feature engineering to achieve satisfying performance, which presented us with the following challenge. Could we combine the advantages of NWP and machine learning to make a more efficient and effective solution? At the same time, single-value (i.e. point estimation) forecasting lacks credibility and flexibility for numerous types of human decision. Could we provide more information to indicate the prediction interval based on high-quality uncertainty quantification?
This paper aims to introduce a unified deep learning method to address these problems through end-to-end learning. In particular, we will predict multiple meteorological variables across different weather stations at multiple  future steps. The proposed approach has several advantages: efficient data pre-processing, end-to-end learning, high accuracy, uncertainty quantification and easy-to-deploy which makes it have considerable practical significance.
The contributions of this work are summarized as follows:
\begin{enumerate}
\item It proposes an effective deep model and information fusion mechanism to handle weather forecasting problems. To the best of our knowledge, this is the first machine learning method which combines historical observations and NWP for weather forecasting. Data and source codes will be released and can be used as a benchmark for researchers to study machine learning in the meteorology field. 

\item It establishes effective assumptions and constructs a novel negative log-likelihood error (NLE) loss function. Unlike Bayesian deep learning (BDL), deep uncertainty quantification (DUQ) can be seamlessly integrated with current deep learning frameworks such as Tensorflow and Pytorch. It can be directly optimized via backpropagation (BP). Our experiments show that compared with typical mean squared error (MSE) and mean absolute error (MAE) loss, training by  NLE loss significantly improves the generalization of point estimation. This phenomenon has never been reported in previous researches.

\item Besides precise point estimation, DUQ  simultaneously inferences the sequential prediction interval. 
This attractive feature has not been studied well in previous deep learning research for time series forecasting. It can be applied to various time series regression scenarios.



\item It explores efficient deep ensemble strategies. The experimental results demonstrate that the ensemble solution significantly improves accuracy. 

\end{enumerate}

The rest of the paper is structured as follows. We discuss related works in Section \uppercase\expandafter{\romannumeral2} and  introduce our method in Section \uppercase\expandafter{\romannumeral3}. In Section \uppercase\expandafter{\romannumeral4}, we discuss experiments and performance analysis
. Last, we conclude with a brief summary and shed light on valuable future works in Section \uppercase\expandafter{\romannumeral6}.

\section{Related Works}
\textbf{Weather Forecasting}
Weather forecasting has been well studied for more than a century. Most contemporary weather forecasting relies on the use of NWP approaches to simulate weather systems using numerical methods \cite{tolstykh2005some, marchuk2012numerical, richardson2007weather}. Some researchers have addressed weather forecasting as a purely data-driven task using ARIMA \cite{chen2011comparison}, SVM \cite{sapankevych2009time}, forward neural network \cite{voyant2012numerical}, etc. These shallow models explore only a few variables, which may not capture the spatio-temporal dynamics of diverse meteorological variables. Deep learning has also shown promise in the field of weather prediction. The study in \cite{hernandez2016rainfall} first adopted an auto-encoder to reduce and capture non-linear relationships between variables, and then trained a multi-layer perceptron for prediction. In \cite{grover2015deep}, a deep hybrid model was proposed to jointly predict the statistics of a set of weather-related variables. The study in \cite{xingjian2015convolutional} formulated precipitation nowcasting as a spatio-temporal sequence forecasting problem and proposed convolutional LSTM to handle it. However, these purely data-driven models  
are limited in that: 1) they all ignore important prior knowledge contained in NWP, which may not capture the spatio-temporal dynamics of diverse meteorological variables; 2) some need tedious feature engineering, such as extracting seasonal features as inputs and kernel selection, which seems contrary to the end-to-end philosophy of deep learning; 3) all lack the flexibility of uncertainty quantification.
\\
\textbf{Deep Learning}
Although deep learning for regression has achieved great success and benefits from the powerful capability of learning representation, solutions like \cite{zhang2018predicting, ZhouMatteson2015, yi2018deep} only focus on point estimation and there is a substantial gap between deep learning and uncertainty quantification.
\\
\textbf{Uncertainty Quantification}
For ease of explaining uncertainty in regression scenario, let us only consider the equation: $\hat{\textbf{Y}}=f(\textbf{X})+\epsilon$, 
where statistically $f(\cdot)$ is the mean estimation (predictable point estimation) of the learned machine learning model and is also called the epistemic part. Its uncertainty comes from \textit{model variance} denoted by $\sigma^2_m$; $\epsilon$ is the irreducible noise, also named the aleatoric part. The reason it exists is because there are unobtained explanatory variables or unavoidable random factors, so it is called \textit{data variance}. Due to the difficulty of expressing $\epsilon$ with a deterministic equation, data variance is usually modeled by a Gaussian distribution with zero mean and a variance $\sigma^2_d$ (Central Limit Theorems). If $\sigma^2$ does not change, it is a homoskedastic problem, otherwise it is regarded as heteroskedastic. Then the total variance $\bm{\sigma}^2=\sigma_d^2 + \sigma_m^2$. The learning process is usually implemented by maximum likelihood estimation, which will learn the estimated $\bm{\hat{\sigma}}^2$.
\\
Uncertainty quantification can provide more reference information for decision-making and has received increased attention from researchers in recent years \cite{khosravi2011comprehensive} . However, most uncertainty quantification methods are based on shallow models and do not take advantages of deep learning. Deep models can automatically extract desirable representations, which is very promising for high-quality uncertainty quantification. To this end, Bayesian deep learning (BDL), which learns a distribution over weights, is currently the most popular technique \cite{wang2016towards}. Nevertheless, BDL has a prominent drawback in that it requires significant modification, adopting variational inference (VI) instead of back-propagation (BP), to train deep models. Consequently, BDL is often more difficult to implement and computationally slower. An alternative solution is to incorporate uncertainty directly into the loss function and directly optimize neural networks by BP \cite{nix1994estimating, lakshminarayanan2017simple, pearce2018high}. This still suffers from certain limitations. 1) Regression is solved as a \textit{mapping} problem rather than \textit{curve fitting}, hence this method cannot naturally be applied to multi-step time series forecasting \cite{roberts2013gaussian, pearce2018high}. 2) The output only consider a single dimension. If it is to be extended to multiple dimensions or for multi-step time series forecasting, the method must be based on effective and reasonable assumptions. 3) Only a shallow forward neural network is used for illustration and the superior performance of deep learning is not explored.
\\
The proposed DUQ addresses these limitations by combining deep learning and uncertainty quantification to forecast multi-step meteorological time series. It can quantify uncertainty, fuse multi-source information, implement multi-out prediction, and can take  advantage of deep models. Meantime, it is optimized directly by BP. 

\section{OUR METHOD}

\subsection{Problem Statement}
Let us say we have historical meteorological observations from a chosen number of weather stations and a preliminary weather forecast from NWP. 
For each weather station, we concern weather forecasting to approximate ground truth in the future. We define this formally below:

\subsubsection{Notations}

For a weather station $s$, we are given:
    \begin{enumerate}
    
        \item Historical observed meteorological time series \\
        $\textbf{E}(t)=[e_1(t),  e_2(t), ... , e_{N_{1}}(t)] \in \mathbb{R}^{N_1}$, where the variable $e_i$ is one type of meteorological element, for $t=1,..., T_{E}$. 
        
        \item Another feature series consist of forecasting timesteps, station ID and NWP forecasting, i.e.,\\ $\textbf{D}(t)=[d_1(t), d_2(t),..., d_{N_2}(t)] \in \mathbb{R}^{N_2}$, where the variable $d_i(t)$ is one of $N_2$ features, for $t=T_{E}+1,..., T_{E}+T_{D}$, and $T_{D}$ is the required number of forecasting steps.
        
        \item Ground truth of target meteorological variables denoted as $\textbf{Y}(t)=[y_1(t), y_2(t),..., y_{N_3}(t)]\in \mathbb{R}^{N_3}$, where the variable $y_i(t)$ is one of $N_3$ target variables, for $t=T_{E}+1, T_{E}+2, ..., T_{E}+T_{D}$ and its estimation denoted as $\hat{\textbf{Y}}(t)$. 
        
        \item Then we define:\\
        $\textbf{E}_{T_E}= [\textbf{E}(1), \textbf{E}(2), ..., \textbf{E}(T_{E})] \in \mathbb{R}^{T_{E} \times N_1}$\\
        $\textbf{D}_{T_D}=[\textbf{D}(T_E+1), \textbf{D}(T_{E}+2),..., \textbf{D}(T_{E}+T_{D})] \in \mathbb{R}^{T_{D} \times N_2}$\\
        $\textbf{X}_{T_{D}}=[\textbf{E}_{T_{E}}; \textbf{D}_{T_{D}}]$\\
        $\textbf{Y}_{T_{D}}= [\textbf{Y}(T_{E}+1), \textbf{Y}(T_{E}+2),...,\textbf{Y}(T_{E}+T_{D}) \in \mathbb{R}^{T_{D} \times N_3}$.  
    \end{enumerate}
    
\subsubsection{Task Definition}

Given $\textbf{X}_{T_{D}}$, the point estimation will predict $\hat{\textbf{Y}}_{T_{D}}$ to approximate $\textbf{Y}_{T_{D}}$ as far as possible. The prediction interval $[\hat{\textbf{Y}}^L_{T_{D}}, \hat{\textbf{Y}}^U_{T_{D}}]$ will ensure ${\textbf{Y}}_{T_{D}} \in [\hat{\textbf{Y}}^L_{T_{D}}, \hat{\textbf{Y}}^U_{T_{D}}]$ (element-wise) with the predefined tolerance probability. The prediction interval will cover the ground truth with at least the expected tolerance probability. 

This research was driven by a real-world weather forecasting competition. For feasible comparison, it focuses on a set time period, i.e., from 3:00 intraday to 15:00 (UTC) of the next day, hence $T_{D}=37$. The target variables include temperature at 2 meters (t2m), relative humidity at 2 meters (rh2m) and wind at 10 meters (w10m), hence $N_3=3$.
The proposed method can be easily extended for any time interval prediction and more target variables.

\begin{figure*}[!htb]
    \centering
        \begin{subfigure}{0.3\textwidth}
              \includegraphics[width=\linewidth]{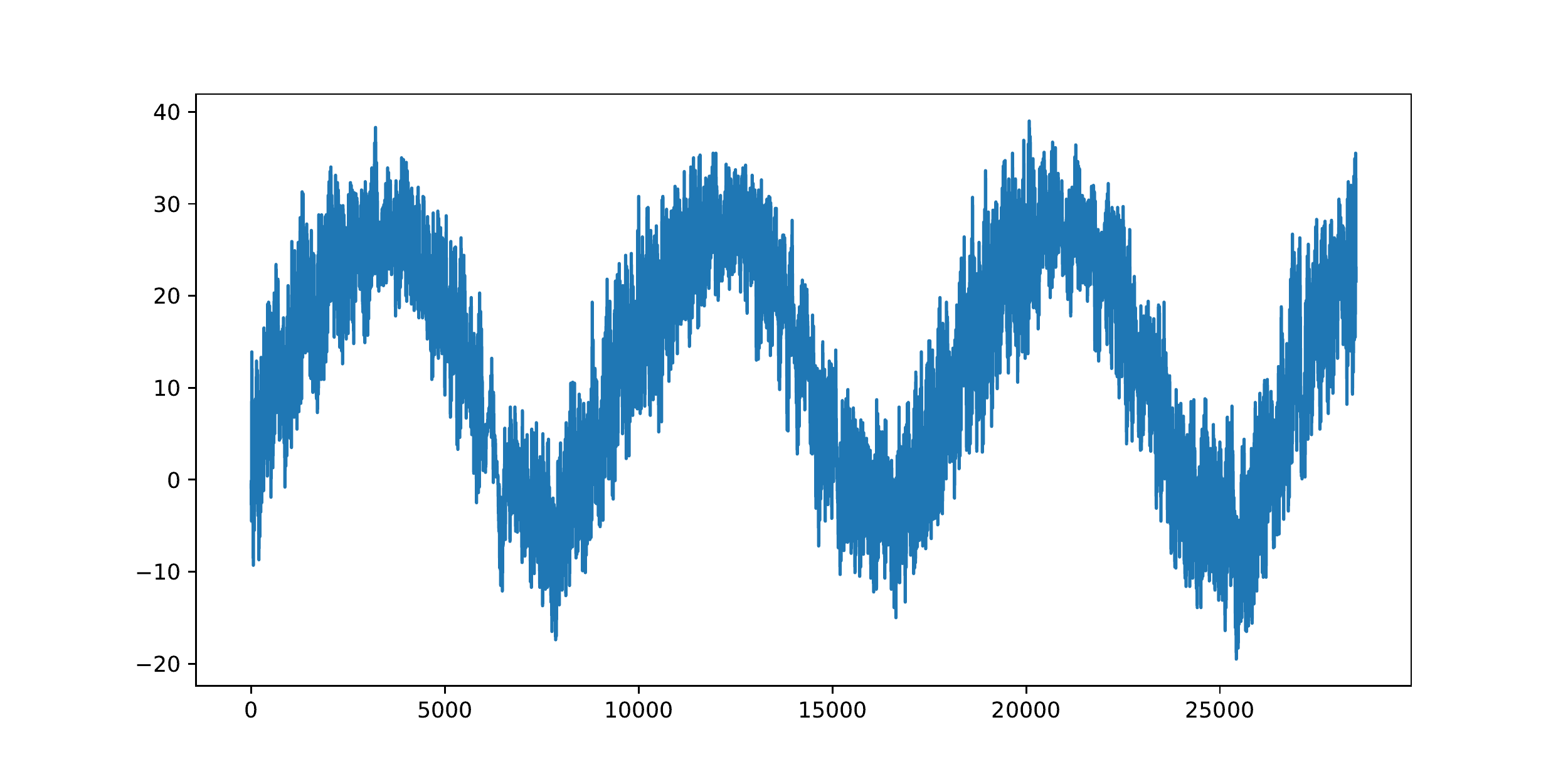}
              \caption{t2m}
              \label{fig:t2m}
        \end{subfigure}
        \begin{subfigure}{0.3\textwidth}
              \includegraphics[width=\linewidth]{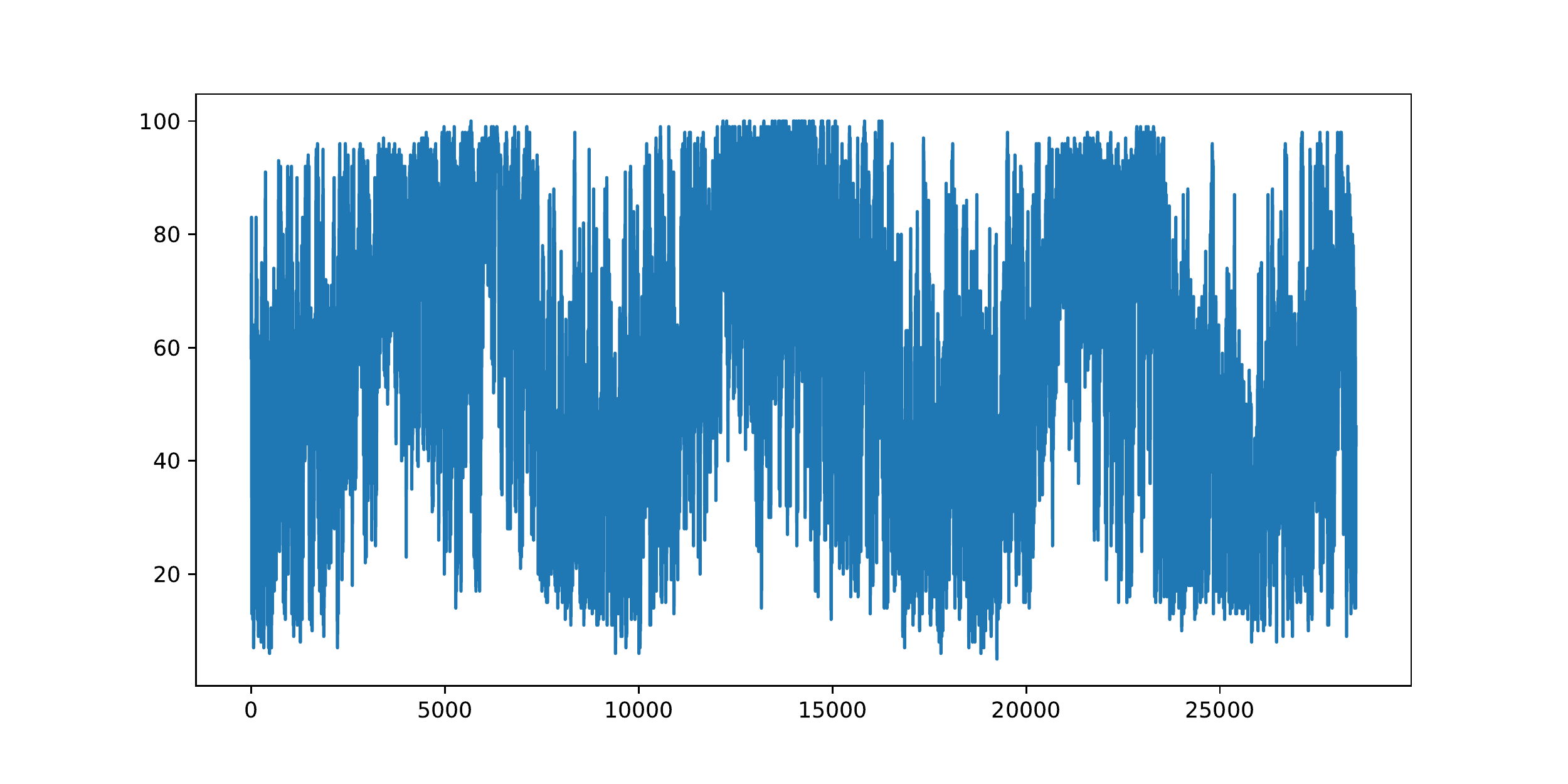}
              \caption{rh2m}
              \label{fig:rh2m}
        \end{subfigure}
        \begin{subfigure}{0.3\textwidth}
              \includegraphics[width=\linewidth]{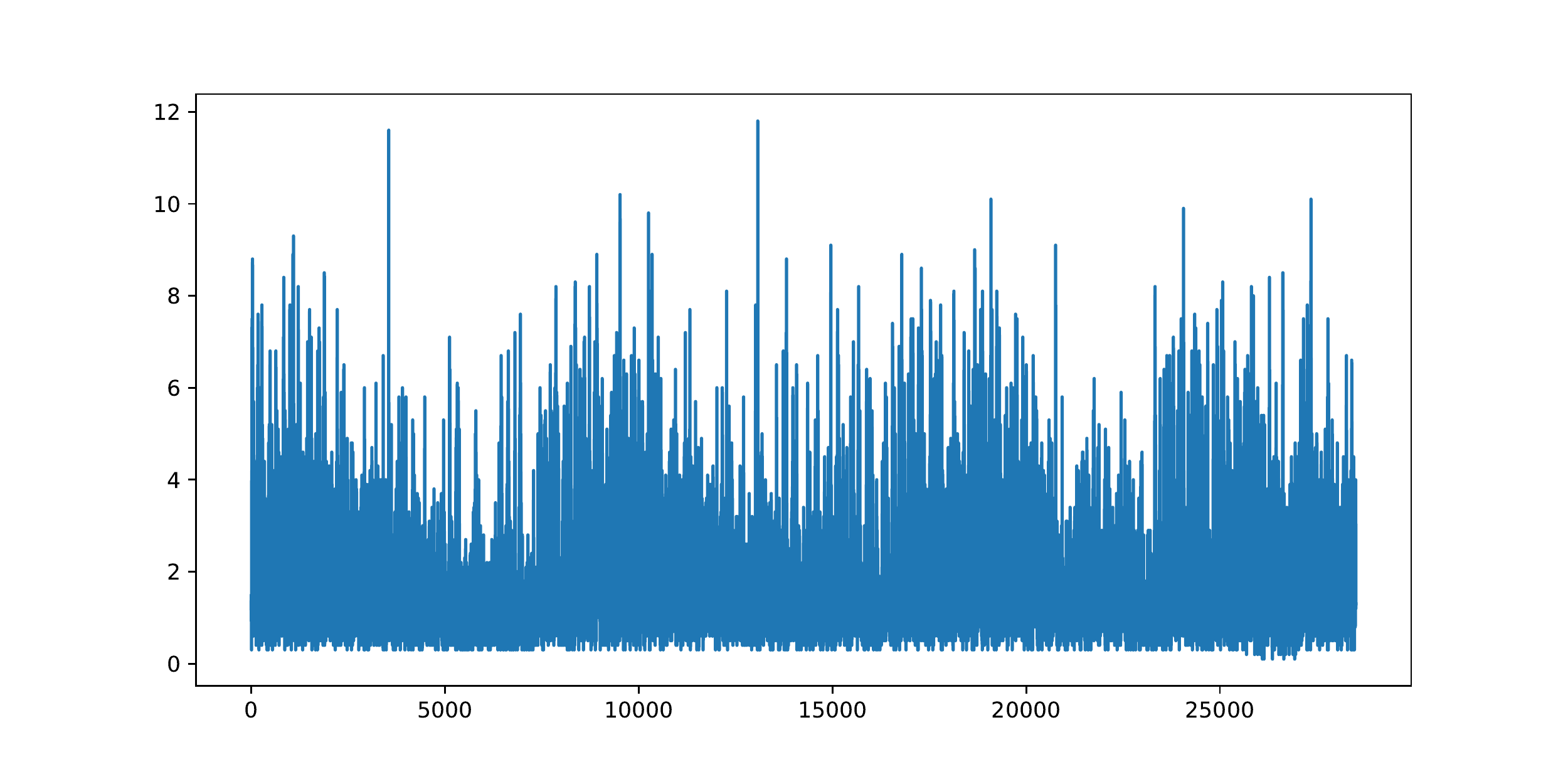}
              \caption{w10m}
              \label{fig:w10m}
        \end{subfigure}
       \caption{Three historical target variable series from 03/01/2015-05/31/2018. They show a strong seasonal variation of t2m, a weak seasonal variation of 2-meter rh2m , and almost no seasonal variation of w10m.}
       \label{fig:hisitory_series_vis}
\end{figure*}
\begin{figure*}[!htb]
    \centering
        \begin{subfigure}{0.3\textwidth}
          \includegraphics[width=\textwidth]{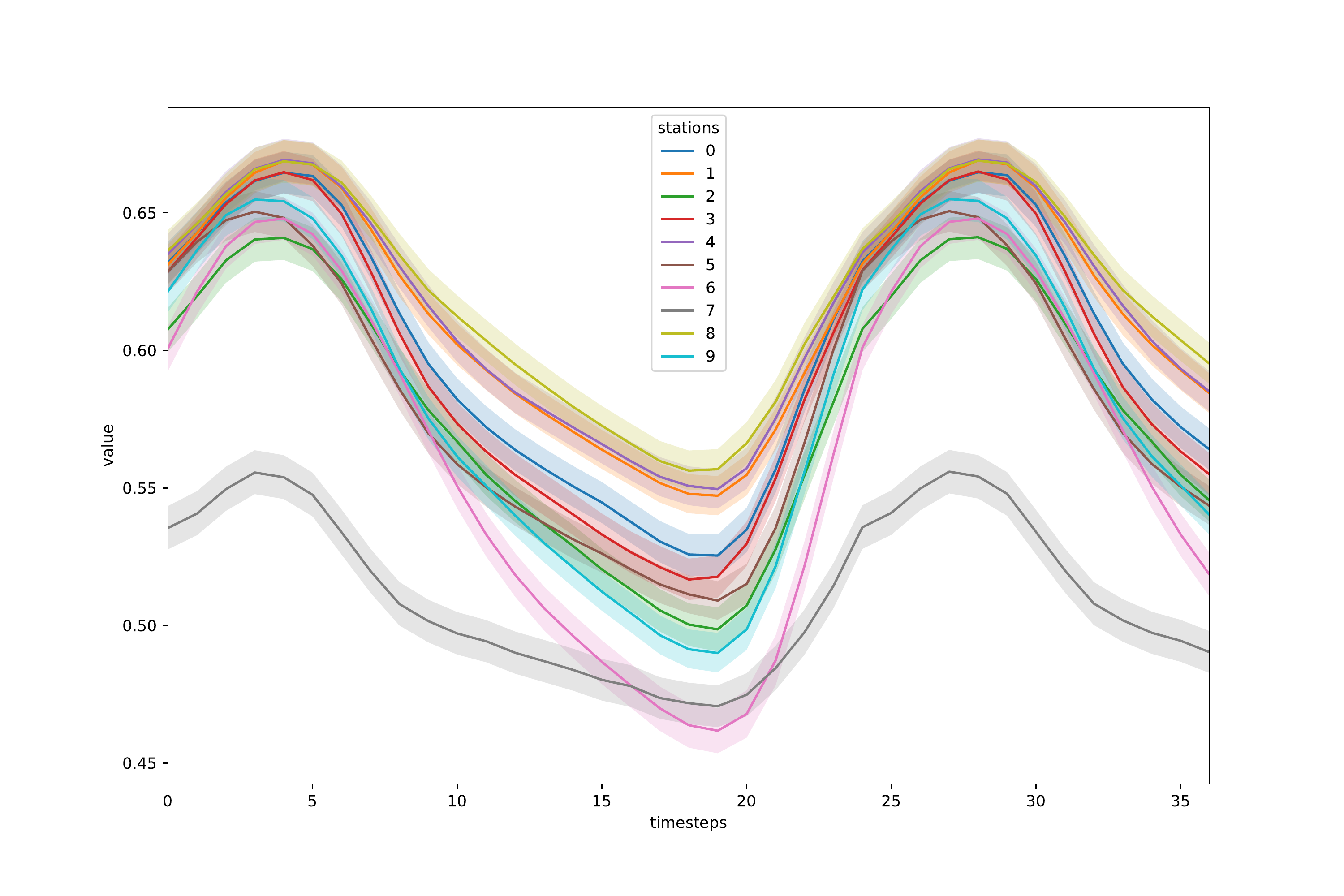}
          \caption{t2m}
          \label{fig:t2m_mv}
        \end{subfigure}
        ~
        \begin{subfigure}{0.3\textwidth}
          \includegraphics[width=\textwidth]{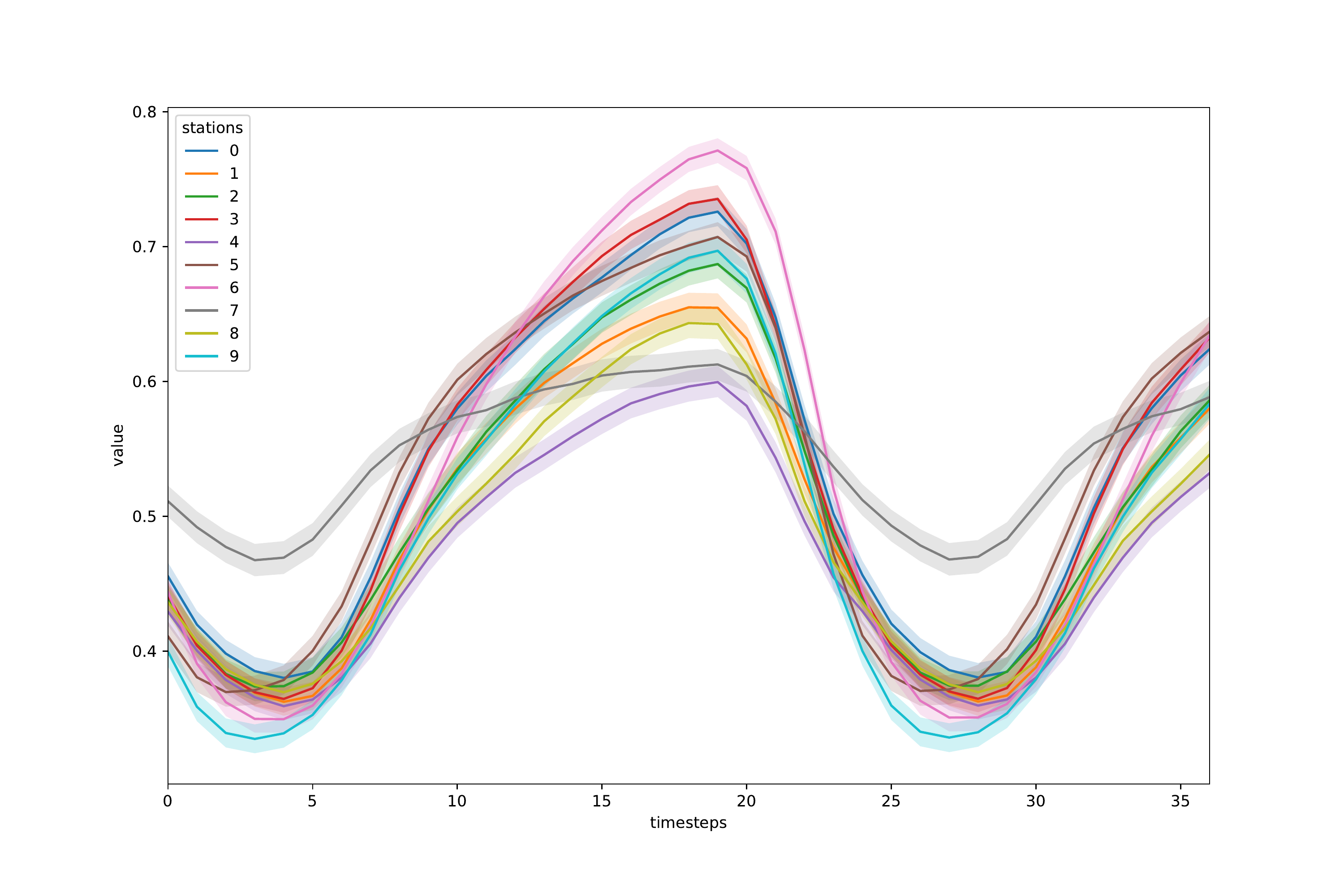}
          \caption{rh2m}
          \label{fig:rh2m_mv}
        \end{subfigure}
        ~
       \begin{subfigure}{0.3\textwidth}
          \includegraphics[width=\textwidth]{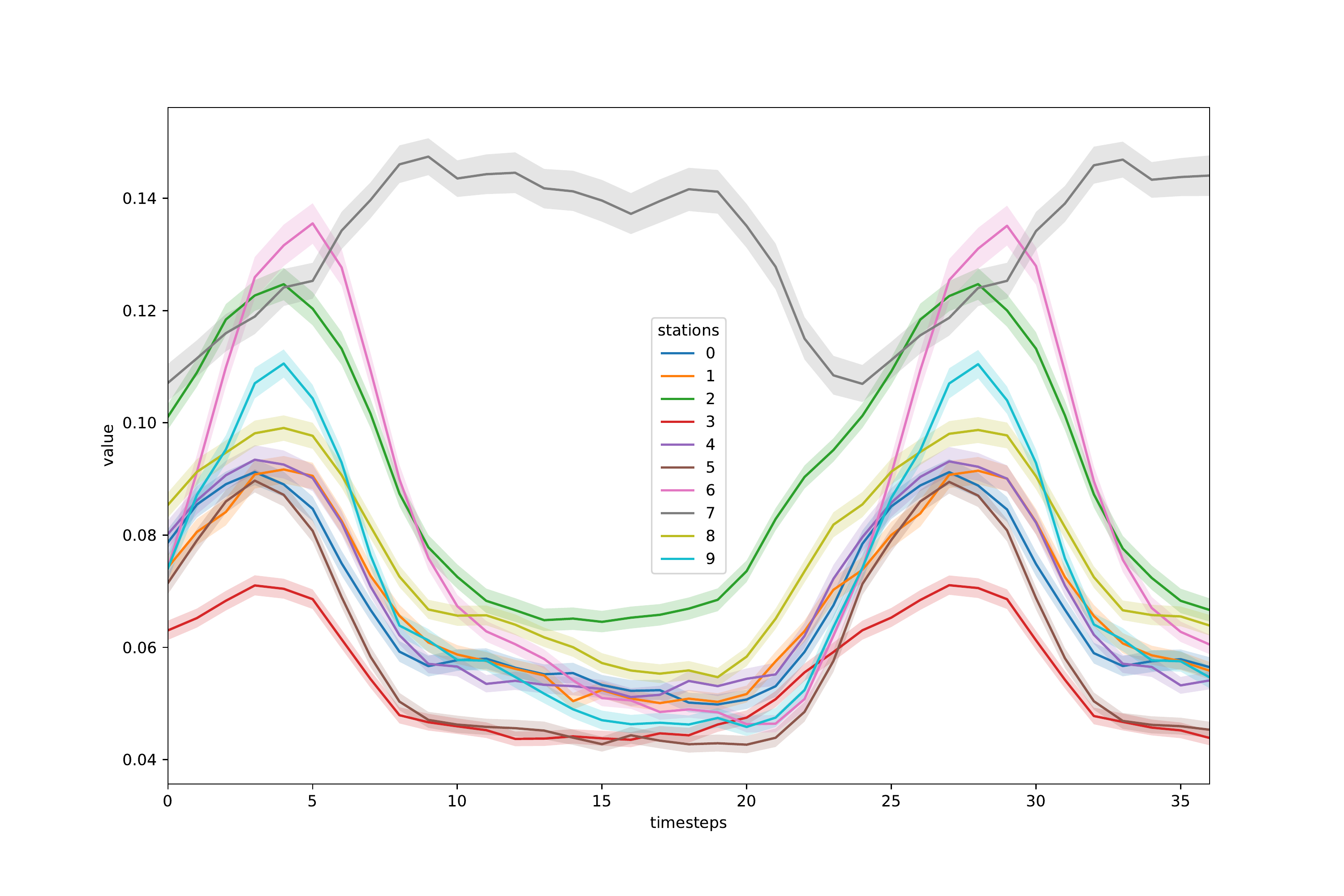}
          \caption{w10m}
          \label{fig:w10m_mv}
       \end{subfigure}
       \caption{The variation of mean (solid line) and 90\% confidence interval (shaded area) for 10 stations during the target forecasting time zone (3:00 intraday to 15:00 of the next day) from 03/01/2015-05/31/2018. Values of Y-axis are following normalization.}
       
       \label{fig:mv}
\end{figure*}

\subsection{Information Fusion Methodology}
\begin{figure*}[!thb]
      \centering
      \includegraphics[width=\textwidth]{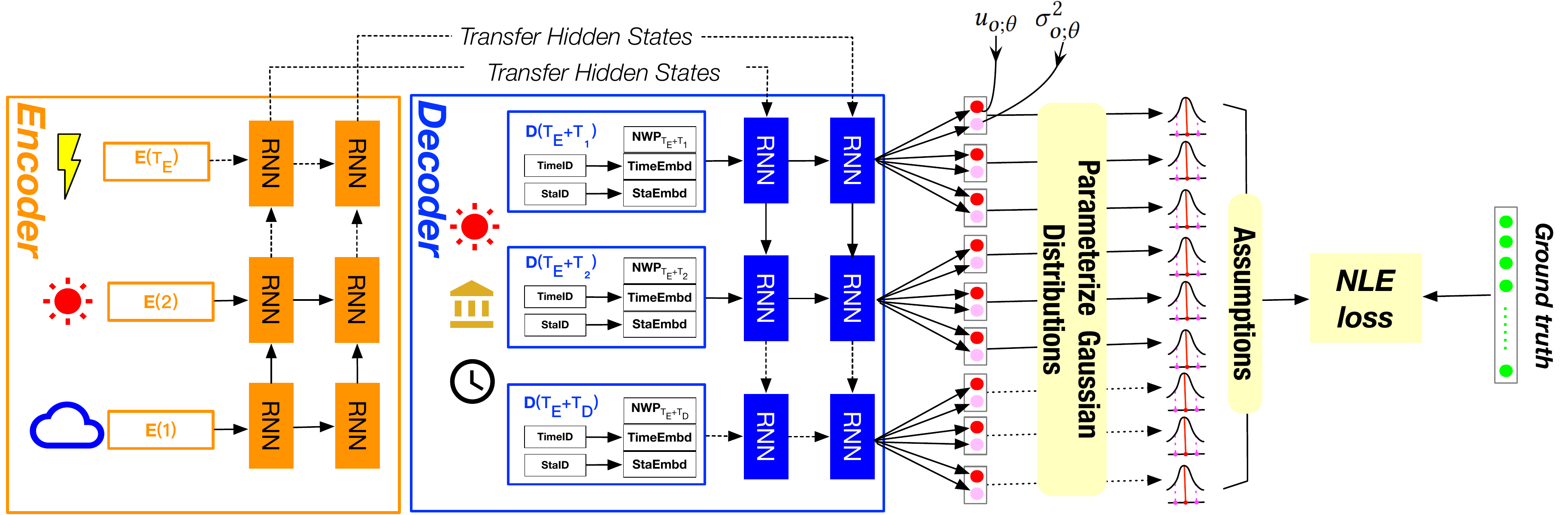}
      \caption{DUQ for sequential point estimation and prediction interval.}
      \label{fig:model_structure}
   \end{figure*}

Data exploration analysis provides insights for the motivation and methodology of information fusion. Fig. \ref{fig:hisitory_series_vis} shows the variation of three target meteorological variables over the past three years. It can seen that only temperature reflects a strong seasonal variation, while relative humidity and wind speed are subjected to much noise. Based on this observation, methods that extract seasonal features from historical data may not provide the best results, since weather changes too dramatically \cite{chen2011comparison,grover2015deep}. Frequent concept drift cause long-term historical meteorological data lack value \cite{lu2018learning}. One conclusion summarizes that \textit{"For many time series tasks only a few recent time steps are required"}\cite{gers2002applying}. On the other hand, NWP is a relatively reliable forecasting method, but inappropriate initial states can introduce undesirable error bias. To address this, we propose a balanced fusion methodology:
\begin{itemize}
    \item First, \textit{only recent} observations, i.e., $\textbf{E}_{T_E}$ should be adopted for \textit{modeling recent meteorological dynamics}.
    \item Second, a wise NWP fusion strategy should incorporate NWP forecasting 
    \textit{at a counterpart forecasting timestep} to easily correcting bias in NWP. Conversely, an unwise fusion strategy that is not carefully designed may absorb NWP  which is not conducive to capturing important NWP signals. Hence we incorporate NWP forcasting into $\textbf{D}_{T_D}$ rather than into $\textbf{E}_{T_E}$ or hidden coding (see Fig. \ref{fig:model_structure}).
\end{itemize}

Fig. \ref{fig:mv} aggregates historical statistics of mean (solid line) and 90\% confidence interval (shade area) for 10 stations from 3:00 intraday to 15:00 (UTC). We find that: 1) There exists obvious difference of mean and variance statistics, e.g., the mean value of station-ID 7 follows a different trend compared with other stations. 2) Every hour at every station has different meteorological characteristics of mean and variance. To address this, we will introduce station ID and time ID into $\textbf{D}_{T_D}$.


\subsection{Data Preprocessing}
\subsubsection{Missing values} There are two kind of missing values, i.e. \textit{block missing} (one-day data lost) and \textit{local missing} (local non-continuous time series), which vary in severity. For block missing \cite{yi2016st}, we just delete the data of those days from the dataset. For local missing data, we use linear interpolation to impute missing values. Taking the training set as an example, we delete 40 days with block missing values from a total of 1188 days, leaving the training data from 1148 (1188-40) days.  
\subsubsection{Normalization of Continuous Variables} Continuous variables without normalization sometimes result in training failure for deep leaning, so we use min-max normalization to normalize each continuous feature into [0, 1]. In the evaluation, we re-normalize the predicted values back to the normal scale.
\subsubsection{Category Variables} There are two category variables, i.e. \textit{Timesteps ID} and \textit{Station ID}. Rather than hard-coding, such as one-hot or sin-cosine coding, we code them by embedding, which has achieved better performance than hard-coding \cite{mikolov2013distributed}.
\subsubsection{Input/Output Tensors} 
Lastly, we load data from all stations and dates and reshape it to three tensors as follows: 
\begin{itemize}
    \item $(I,T_E,S,N_1), (I,T_D,S,N_2)$ (i.e., input tensors).
    \item $(I,T_D,S,N_3)$ (i.e., ground truth tensor).
\end{itemize} 
Note that $I$ is the date index and $S$ is the station index. When drawing training samples, we first draw integer date $i \in I$ and station $s \in S$. We can then index by $i, s$ from these three tensors and obtain one training instance $\textbf{X}_{T_{D}}^{i,s}$=$[\textbf{E}_{T_{E}}^{i,s}; \textbf{D}_{T_{D}}^{i,s}]$ and $\textbf{Y}_{T_D}^{i,s}$ abbreviated as $\textbf{X}_{T_{D}}$=$[\textbf{E}_{T_{E}}; \textbf{D}_{T_{D}}]$ and $\textbf{Y}_{T_D}$ for brevity.
The advantage of organizing data in this four-tuple style is that we can conveniently index the data via the specific dimension for hindsight inspection and consideration of scalability. For example, we can index specific dates and stations for later transfer learning research. Readers can refer to
the instantiated example \textit{Parameter Settings for Reproducibility} in Section \uppercase\expandafter{\romannumeral5} for deeper understanding.

\subsection{Model Architecture}
The proposed DUQ is based on sequence-to-sequence (seq2seq, also a.k.a Encoder-Decoder). Its detailed formula is not discussed here. Readers can refer to \cite{srivastava2015unsupervised} for more detail. There are already many high-performance variants for different tasks, but most of them focus on making improvements from the structural perspective to make point estimation more precise. We first incorporate sequential uncertainty quantification for weather forecasting into the architecture presented in Fig. \ref{fig:model_structure}. The encoder first extracts latent representations $\textbf{c}$ from the observed feature series $\textbf{E}_{T_{E}}$:
\begin{displaymath}
    \textbf{c} = Enc(\textbf{E}_{T_{E}}; \theta_1)
\end{displaymath}
where $\textbf{c}$ captures the current meteorological dynamics and is then transferred to form the initial state of the decoder. Based on the memory of $\textbf{c}$, the decoder absorbs $\textbf{D}_{T_{D}}$ including station identity (StaID), forecasting time identity (TimeID), and NWP forecasting. Two embedding layers will be introduced for StaID and TimeID respectively to automatically learn the embedding representations. This architecture will generate sequential point estimation $\textbf{u}_{T_{D}}$ used as $\hat{\textbf{Y}}_{T_{D}}$  to predict ${\textbf{Y}}_{T_{D}}$ as well as the variance $\bm{\hat{\sigma}}^2_{T_{D}}$ utilized to estimate $[\hat{\textbf{Y}}^L_{T_{D}}, \hat{\textbf{Y}}^U_{T_{D}}]$:
 
\begin{displaymath}
    \bm{\hat{\sigma}}^2_{T_{D}}, \hat{\textbf{Y}}_{T_{D}}=Dec(\textbf{c}, \textbf{D}_{T_{D}}; \theta_2)
\end{displaymath}
where $\theta_1$ and $\theta_2$ are learnable parameters. We use $f(\cdot)$ to represent the combination of Encoder-Decoder and use $\textbf{X}_{T_{D}}=[\textbf{E}_{T_{E}}; \textbf{D}_{T_{D}}]$ which can then be regarded as:

\begin{displaymath}
\bm{\hat{\sigma}}^2_{T_{D}}, \hat{\textbf{Y}}_{T_{D}}=f(\textbf{X}_{T_{D}})
\end{displaymath}

\subsection{Learning Phase}
DUQ predicts two values at each timestep corresponding to the predicted mean and variance to parameterize the Gaussian distributions \footnote[2]{We enforce the positivity constraint on the variance by passing the second output through the $softplus$
function $log(1 + exp(\cdot)$), and add a minimum variance (e.g. $10^{-6}$) for numerical stability.}. The NLE is calculated for the Gaussian distributions, which must be based on reasonable hypotheses. Three mild but experimentally effective assumptions are proposed (degree of effectiveness can be seen in the experimental results in Table \ref{tab_picp}):
\begin{enumerate}
    \item Each day and each station are independent. This assumption ensures it is reasonable that the number of all training samples can be regard as $I\times S$. Based on this, we can minimize the negative log-likelihood error loss:

    \begin{displaymath}
    \begin{aligned}
        NLE&= -\prod_{i=1}^{I}\prod_{s=1}^{S} p_{\theta}(\textbf{Y}_{T_{D}}|\textbf{X}_{T_{D}})
    \end{aligned}
    \end{displaymath}
    
    \item Each target variable and each timestep at one specific station are \textit{conditionally independent} given $\textbf{X}_{T_{D}}$. Based on this, we can further decompose $p_{\theta}(\textbf{Y}_{T_{D}}|\textbf{X}_{T_{D}})$ by the product rule and transform it via log operation as:

        \begin{displaymath}
            \begin{aligned}
            p_{\theta}(\textbf{Y}_{T_{D}}|\textbf{X}_{T_{D}})&= \prod_{o=1}^{N_3}\prod_{t=1}^{T_{D}} p_{\theta}(y_{o}(t)|\textbf{X}_{t})\\
            \end{aligned}
        \end{displaymath}
        
        \begin{displaymath}
        \begin{aligned}
        log\ p_{\theta}(\textbf{Y}_{T_{D}}^i|\textbf{X}_{T_{D}}^i)&= \sum_{o=1}^{N_3}\sum_{t=1}^{T_{D}}log\  p_{\theta}(y_{o}(t)|\textbf{X}_{t}^i)\\
        \end{aligned}
        \end{displaymath}
    
    \item The target variables satisfy multivariate independent Gaussian distribution and $\sigma_{\theta}$ is a function of the input features, i.e., $\textbf{Y}_{T_{D}} \sim N(u_{\theta}(\textbf{X}_{T_{D}}), \sigma_{\theta}(\textbf{X}_{T_{D}}))$. Based on this assumption, the final loss is:
    \begin{displaymath}
    \begin{aligned}
        NLE&= -\sum_i^I\sum_{o=1}^{N_3}\sum_{t=1}^{T_{D}}log\  p_{\theta}(y_{o}(t)|\textbf{X}_{t}^i)\\
        &= \sum_i^I\sum_{o=1}^{N_3}\sum_{t=1}^{T_{D}} \frac{log\sigma_{o;\theta}^{2}(\textbf{X}_{t}^i)}{2} +\frac{(y_o(t)-u_{o;\theta}(\textbf{X}_{t}^i))^2}{2\sigma_{o;\theta}^{2}(\textbf{X}_{t}^i)} + C\
    \end{aligned}
    \end{displaymath}
\end{enumerate} 
where $C$ is a constant which can be omitted during training. $y_o(t)$ is the ground truth of a target variable $o$ at timestep $t$, $\sigma^2_{o;\theta}(\textbf{X}_{t}^i)$ and $u_{o;\theta}(\textbf{X}_{t}^i)$ are respectively the variance and mean of a Gaussian distribution parameterized by DUQ. The aim of the entire learning phase is to minimize NLE. Optimizing by deep models can easily lead to overfitting on the training set, therefore it is necessary to implement early-stopping on the validation set. Algorithm 1 outlines the procedure for the learning phase.
        \begin{algorithm}
            \caption{Algorithm for learning}
            \label{algr}
            \SetKwInOut{Input}{Input}
            \SetKwInOut{Output}{Output}
        
            \Input{$N_1, N_2, N_3, T_{E}$, $T_{D}$;\\
            Input tensors: $(I,T_E,S,N_1)$, $(I,T_D,S,N_2)$; \\
            Output tensor: $(I,T_D,S,N_3)$; \\
            Maximum iterations; \\
            Tolerance iterations for early-stopping; \\
            \\}
            \Output{Learned DUQ model}
            
            \nonl // Learning phase \\
            
        Initialize all learnable parameters $\theta$ in DUQ

        \Repeat{stopping criteria are met}{
            
            $\mathcal{B} \leftarrow \emptyset$ \\
            
            \While{each training datum n ($1 \le n \le Batch Size$)}{
                \nonl // Format training data samples \\
                Draw a random integer $i \sim uniform(0, I)$\\
                Draw a random integer $s \sim uniform(0, S)$\\
                Index by $i,s$ from input and output tensors and get one training sample ($\textbf{X}_{T_{D}} ,\textbf{Y}_{T_D}$)\\
                Put this sample into $\mathcal{B}$
            }
            Update $\theta$ via BP by minimizing the NLE loss on $\mathcal{B}$
        }
        \end{algorithm}

\subsection{Inference Phase}

After training, we can implement statistical inference for an input $\textbf{X}_{T_{D}}$ by:
\begin{displaymath}
    u_{\theta}(\textbf{X}_{T_{D}}),\sigma^2_\theta(\textbf{X}_{T_{D}})=f(\textbf{X}_{T_{D}})
\end{displaymath}
where $u_{\theta}(\textbf{X}_{T_{D}})$ is statistically the mean estimation i.e., $\hat{\textbf{Y}}_{T_{D}}$ given $\textbf{X}_{T_{D}}$, which will be adopted for forecasting and  $\sigma^2_\theta(\textbf{X}_{T_{D}})$ is statistically the variance estimation, i.e., $\hat{\bm{\sigma}}^2_{T_{D}}$ given $\textbf{X}_{T_{D}}$. Recall our assumption that $\hat{\textbf{Y}}_{T_{D}}$ satisfies Gaussian distribution, so upper bound $\hat{\textbf{Y}}^U_{T_{D}}$and lower bound $\hat{\textbf{Y}}^L_{T_{D}}$ can be inferenced as follows:

\begin{displaymath}
    \hat{\textbf{Y}}^U_{T_{D}}= \hat{\textbf{Y}}_{T_{D}} + \lambda \hat{\bm{\sigma}}_{T_{D}}
\end{displaymath}
\begin{displaymath}    
    \hat{\textbf{Y}}^L_{T_{D}}= \hat{\textbf{Y}}_{T_{D}} - \lambda \hat{\bm{\sigma}}_{T_{D}}
\end{displaymath}
where $\hat{\bm{\sigma}}_{T_{D}}$ is the standard deviation and, $\lambda$ should be determined according to the pre-defined $1-z$. In this research, $1-z=0.9$ thus $\lambda$ is set to 1.65 according to the z-score of Gaussian distribution. Algorithm \ref{algr2} gives the inference procedure. 
        \begin{algorithm}
            \caption{Algorithm for inference}
            \label{algr2}
            \SetKwInOut{Input}{Input}
            \SetKwInOut{Output}{Output}
        
            \Input{$\textbf{X}_{T_{D}}$, $z$;\\}
            \Output{$\hat{\textbf{Y}}^L_{T_{D}}, \hat{\textbf{Y}}^U_{T_{D}}, \hat{\textbf{Y}}_{T_{D}}, \hat{\bm{\sigma}}_{T_{D}}$;}
            
            \nonl // Inference phase \\

            $\hat{\textbf{Y}}_{T_{D}},\hat{\bm{\sigma}}_{T_{D}}=f(\textbf{X}_{T_{D}})$ \\   
            
            \nonl // determine $\lambda$ by $z$ from z-score of Gaussian distribution
            
            $\hat{\textbf{Y}}^L_{T_{D}}= \hat{\textbf{Y}}_{T_{D}} - \lambda \hat{\bm{\sigma}}_{T_{D}}$ \\
            
            $\hat{\textbf{Y}}^U_{T_{D}}= \hat{\textbf{Y}}_{T_{D}} + \lambda \hat{\bm{\sigma}}_{T_{D}}$ \\
            
            de-normalize $\hat{\textbf{Y}}^L_{T_{D}}, \hat{\textbf{Y}}^U_{T_{D}}, \textbf{Y}_{T_{D}}$ for real-world evaluation.
        \end{algorithm}
        
\subsection{Ensemble Methodology}
We adopt a simple but efficient principle for ensemble: each single model is a DUQ-based model initialized with specified nodes. The ensemble point estimation is the averaged point estimation of all DUQ-based models, which is scalable and easily implementable.        
\subsection{Evaluation Metrics}
\subsubsection{Point Estimation Measurement}
We first calculate the root mean squared error ($RMSE$) for each objective variable from all stations for daily evaluation.
\begin{displaymath}
    RMSE_{obj}=\sum_{s=1}^S||\textbf{Y}_{T_D}^{s,obj} - \hat{\textbf{Y}}_{T_D}^{s,obj}||
\end{displaymath}
where $\textbf{Y}_{T_D}^{s,obj}$ and $\hat{\textbf{Y}}_{T_D}^{s,obj}$ are respectively the ground truth and the  predicted value of the objective variable (i.e., $t2m$, $rh2m$ or $w10m$ in this paper) at station $s$.

\begin{displaymath}
RMSE_{day}=\frac{RMSE_{t2m}+RMSE_{rh2m}+RMSE_{w10m}}{N_3}
\end{displaymath}

$RMSE_{day}$ is the ultimate RMSE criterion in the experimental reports for each day. $RMSE_{avg}$ is the average $RMSE_{day}$ over all days. To demonstrate the improvement over the classic NWP method, we employ the following evaluation using the associated skill score ($SS$, the higher the better):
\begin{displaymath}
SS_{obj}= 1- \frac{RMSE_{obj\_ml}}{RMSE_{obj\_nwp}}
\end{displaymath}
where $RMSE_{obj\_nwp}$ is the $RMSE_{obj}$ calculated by the NWP method and $RMSE_{obj\_ml}$ is calculated from the prediction made of machine learning models.

\begin{displaymath}
SS_{day}=\frac{SS_{t2m}+SS_{rh2m}+SS_{w10m}}{N_3}
\end{displaymath}

$SS_{day}$ is the ultimate $SS$ criterion in experimental reports for every day. $SS_{avg}$ is the average $SS_{day}$ over all days, which is also the ultimate rank score in the online competition.

\subsubsection{Prediction Interval Measurement}

To evaluate the prediction interval, we introduce the metric called prediction interval coverage probability (PICP). First, an indicator vector \\ $\textbf{B}(t)=[b_1(t), b_2(t),..., b_{N_3}(t)] \in \mathbb{R}^{N_3}$ is defined, for $t=T_{E}+1,..., T_{E}+T_{D}$. Each Boolean variable $b_o(t) \in [0,1]$ represents whether the objective variable $o$ at the predicted time step $t$ has been captured by the estimated prediction interval.

    \begin{displaymath}
        b_o(t)=
        \begin{cases}
        1, & \text{if \quad} \hat{y}^L_o(t)\leq y_o(t) \leq \hat{y}^U_o(t) \\
        0, & \text{else.}
        \end{cases}
        \label{formula_b}
    \end{displaymath}

The total number of captured data points for the objective variable is defined as $C_{obj}$,
$$
C_{obj} = \sum_{s=1}^{S}\sum_{t=T_{E}+1}^{T_{E}+T_{D}} b_o(t)
$$

Then $PICP_{obj}$ for the objective variable is defined as:
\begin{displaymath}
    PICP_{obj}=\frac{C_{obj}}{T_D*S}
\end{displaymath}

Ideally, $PICP_{obj}$ should be equal to or greater than the pre-defined value, i.e., $1-z=0.9$ where $z$ is the significant level and is set to 0.1 in our experiments.

\section{Experiments and PERFORMANCE ANALYSIS}

\subsection{Baselines}
\textbf{SARIMA} Seasonal autoregressive integrated moving average is a benchmark model for univariate time series, where parameters are chosen using AIC (Akaike information criterion).

\textbf{SVR} Support vector regression is a non-linear support vector machine for regression estimation. 

\textbf{GBRT} Gradient boosting regression
tree is an ensemble method for regression
tasks and is widely used in practice.


\textbf{DUQ$_{50}$} is one layer GRU-based seq2seq with 50 hidden nodes. The loss function is $NLE$.

\textbf{DUQ$_{50-50}$} is two layers GRU-based seq2seq with 50 hidden nodes of each layer. The loss function is $NLE$.

\textbf{DUQ$_{200}$} is one layer GRU-based seq2seq with 200 hidden nodes. The loss function is $NLE$.

\textbf{DUQ$_{300-300}$} is two layers GRU-based seq2seq with 300 hidden nodes of each layer. The loss function is $NLE$.

\textbf{DUQ$_{noNWP}$} is the same as DUQ$_{300-300}$ except that NWP forecasting (i.e. $\textbf{NWP}$ of $\textbf{D}_{T_D}$, refer to Fig. \ref{fig:model_structure} ) is masked by zero values. 

\textbf{DUQ$_{noOBS}$} is the same as DUQ$_{300-300}$ except that the observation features (i.e. $\textbf{E}_{T_E}$) are masked by zero values.

\textbf{Seq2Seq$_{MSE}$} is the same as DUQ$_{300-300}$ except that the loss function is MSE.

\textbf{Seq2Seq$_{MAE}$} is the same as DUQ$_{300-300}$ except that the loss function is MAE.

\textbf{DUQ$_{Esb3}$} ensembles three DUQ models (i.e., DUQ$_{300-300}$, DUQ$_{200-200}$, DUQ$_{100-100}$) for online evaluation. This method achieved 2nd place in the online competition.

\textbf{DUQ$_{Esb10}$} ensembles 10 DUQ models with different architecture to explore the effectiveness of the ensemble. It ensembles DUQ$_{300-300}$, DUQ$_{310-310}$, ..., DUQ$_{390-390}$ (increasing at 10-neuron intervals).

\textbf{Model$_{1st}$} achieves the best $SS_{avg}$ during online comparison. According to the on-site report, the author also adopted a complicated stacking and ensemble learning strategy.

\subsection{Experimental Environments} The experiments were implemented on a GPU server with Quadro P4000 GPU and Keras programming environment (Tensorflow backend).

\subsection{Parameter Settings for Reproducibility}
The \textit{batch size} is set to 512. The \textit{embedding dimension} of each embedding layer is set to 2. Since we adopted an early-stopping strategy, it was not necessary to set the \textit{epoch} parameter. Instead, we set the number of \textit{maximum iterations} to a relatively large number of 10000 to take sufficient batch iterations into consideration. 
The \textit{validation interval} (vi) is set to 50 meaning that for every 50 iterations, we will test our model on validation set and calculate the validation loss. We set the \textit{early-stopping tolerance} (est) to 10, meaning that if the validation loss over 10 continuous iterations did not decrease, training would be stopped early. We defined the \textit{validation times} (vt) when early-stopping was triggered, hence the \textit{total iterations} (ti) can be calculated by ti=vt$\times$vi. For the prediction interval, $z $ was set to 0.1, $1-z=0.9$ thus $\lambda$ is set to 1.65 according to the z-score of Gaussian distribution.

We set $N_1=9, N_2=31, N_3=3, T_{E}=28, T_{D}=37$ to preprocess the original dataset. After preprocessing, the final dataset shape was as shown below.

For the training set:

\begin{itemize}
\item Encoder inputs: (1148, 28, 10, 9)
\item Decoder inputs: (1148, 37, 10, 31)
\item Decoder outputs: (1148, 37, 10, 3)
\end{itemize}

For the validation set:

\begin{itemize}
\item Encoder inputs: (87, 28, 10, 9)
\item Decoder inputs: (87, 37, 10, 31)
\item Decoder outputs: (87, 37, 10, 3)
\end{itemize}

For the test set on each day:
\begin{itemize}
\item Encoder inputs: (1, 28, 10, 9)
\item Decoder inputs: (1, 37, 10, 31)
\item Decoder outputs: (1, 37, 10, 3)
\end{itemize}

The meaning of each number is explained as follows: we acquired data from 1148 days for the training set and data from 87 days for the validation set. Because our evaluation is based on online daily forecasting, the test day index is 1. Number 28 is a hyperparameter, meaning that the previous 28 hours of observations were used to model recent meteorological dynamics. Number 37 was set according to the specified forecasting steps for the next 37 hours. Number 9 is the dimension of observed meteorological variables. Number 31 (dimension of decoder inputs) consists of concatenating \textit{Timesteps ID} and \textit{Station ID} into 29-dimension of NWP forecasting (2+29=31). Number 3 is the ground truth number for 3 target variables. The size of the final training set is 1148*10=11480. The size of validation set is 87*10=870, which is used for early-stopping. The size of test set on each day is 1*10=10.

\subsection{Performance analysis}
Table \ref{table_SS} presents the evaluation by $SS$ score based on rolling forecasting, with incremental data releasd on a daily basis for nine days to mimic real-world forecasting processes.

\subsubsection{Effect of information fusion}
Comparing DUQ$_{300-300}$ \\ with DUQ$_{noNWP}$  validates the effectiveness of fusing NWP forecasting. Comparing DUQ$_{300-300}$ with DUQ$_{noOBS}$ validates the effectiveness of modeling recent meteorological dynamics. 

\subsubsection{Effect of deep learning}
On average, the deep learning-based models (DUQ and Seq2Seq) perform better than the non-deep learning models (SARIMA, SVR, GBRT). Comparing DUQ$_{50}$ and DUQ$_{50-50}$ validates the influence of deeper layers. Comparing DUQ$_{50}$, DUQ$_{200}$, and DUQ$_{300-300}$ validates the effectiveness of nodes under the same number of layers. 

\subsubsection{Effect of loss function} A notable result is that DUQ$_{300-300}$ trained by NLE loss performs much better than Seq2Seq$_{MSE}$ (MSE loss) and Seq2Seq$_{MAE}$ (MAE loss). In order to empirically  understand the reasons for better generalization when trained by NLE, we calculated the $ti$ when early-stopping was triggered, as shown in Table \ref{tab_iter}. It can be seen that DUQ$_{300-300}$ requires more iterations to converge. A reasonable interpretation is that NLE loss jointly implements two tasks i.e., mean optimization and variance optimization, which need more iterations to converge. This joint optimization may to some extent play a regularization role and help each other out of the local minimum. It may therefore require more iterations to converge and may have better generalization. We believe that this phenomenon deserves the attention of researchers, and that it should be proved by theory in follow-up work.

\subsubsection{Effect of ensemble}
The ensemble model DUQ$_{Esb3}$ was used in the online competition \footnote[3]{Readers can refer to https://challenger.ai/competition/wf2018 to check our online scores which are consistent with DUQ$_{Esb3}$ during Day 3-Day 9. During Day 1 and Day 2 of the online competition, DUQ$_{Esb3}$ had not been developed. In this paper we re-evaluate DUQ$_{Esb3}$ offline on Day 1 and Day 2. This is also why DUQ$_{Esb3}$ with SS$_{avg}$=0.4673 is better than the Model$_{1st}$ (0.4671) but was only awarded 2nd place online.}. DUQ$_{Esb10}$ achieved the best $SS_{avg}$, which indicates that ensemble with more DUQ models would provide a better solution. 

\subsubsection{Significance of T-test}
Because real-world forecasting only covers nine days, we implemented a \textit{one-tail paired T-test} with significance level $sig=0.25$ to ensure that our results were statistically significant. The column \textit{P-value} between DUQ$_{Esb10}$ and others shows that each T-test has been passed which means that our method DUQ$_{Esb10}$ is significantly better than any other baseline under the specified significance level. For the single model, we also implemented a T-test between DUQ$_{300-300}$ and other baselines including DUQ$_{noOBS}$, DUQ$_{noNWP}$, Seq2Seq$_{MSE}$, Seq2Seq$_{MAE}$ to ensure that DUQ$_{300-300}$ had significant effectiveness, which is shown in Table. \ref{tab_p_value}.

\subsubsection{Evaluation by RMSE}
We also evaluated all methods by $RMSE_{avg}$ as shown in Table \ref{table_RMSE}. Since $RMSE_{avg}$ and $SS_{avg}$ do not have a fully linear relationship, the counterpart assessment does not reach the optimum at the same time while DUQ$_{Esb10}$ still achieves the best $RMSE_{avg}$. The related P-value also indicates that DUQ$_{Esb10}$ is significantly better than other baselines under $sig=0.25$. Because online evaluation does not release the RMSE ranking, the RMSE of Model$_{1st}$ place is not shown in Table \ref{table_RMSE}.

\subsubsection{Discuss instability of weather forecasting}
Due to meteorological instability and variability,  no single model can achieve the best scores every day - not even the ensemble method. Sometimes a single model can achieve the highest $SS_{day}$ score, such as DUQ$_{200}$ on Day 2 and DUQ$_{50-50}$ on Day 7. Overall, however, the ensemble method DUQ$_{Esb10}$ achieves the greatest score benefit from the stability of ensemble learning. The instability of meteorological elements also reflects the need for a prediction interval.

\subsubsection{Quantity of prediction interval}
An effective prediction interval should satisfy that $PICP_{obj}$ is equal to or greater than the pre-defined $1-z=90\%$. Table \ref{tab_picp} shows the results. In particular, the $PICP_{rh2m}$ on Day 3 seems far below expectations. The main reason is that forecasting of $rh2m$ is not accurate on that day. The online competition scores of all contestants were particular low on that day. Generally, our approach  meets the requirement $PICP_{avg} \geq 1-z=90\%$.

\subsubsection{Quality of prediction interval}
\begin{figure*}[!htb]
    \centering
        \begin{subfigure}{0.3\textwidth}
          \includegraphics[width=\textwidth]{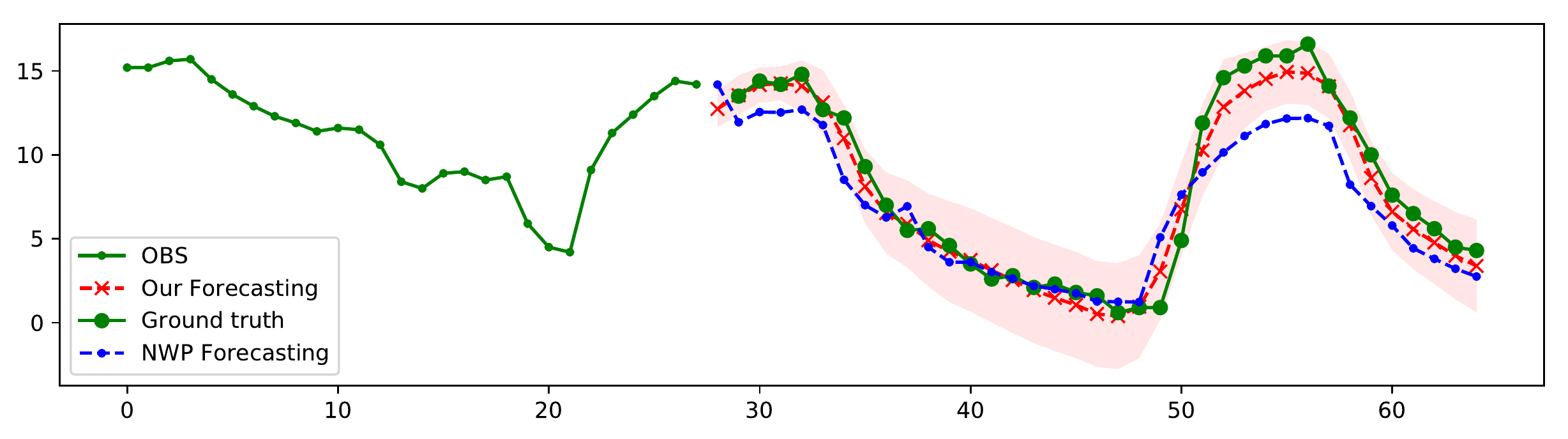}
          \caption{t2m}
          \label{fig:1}
        \end{subfigure}
        ~
        \begin{subfigure}{0.3\textwidth}
          \includegraphics[width=\textwidth]{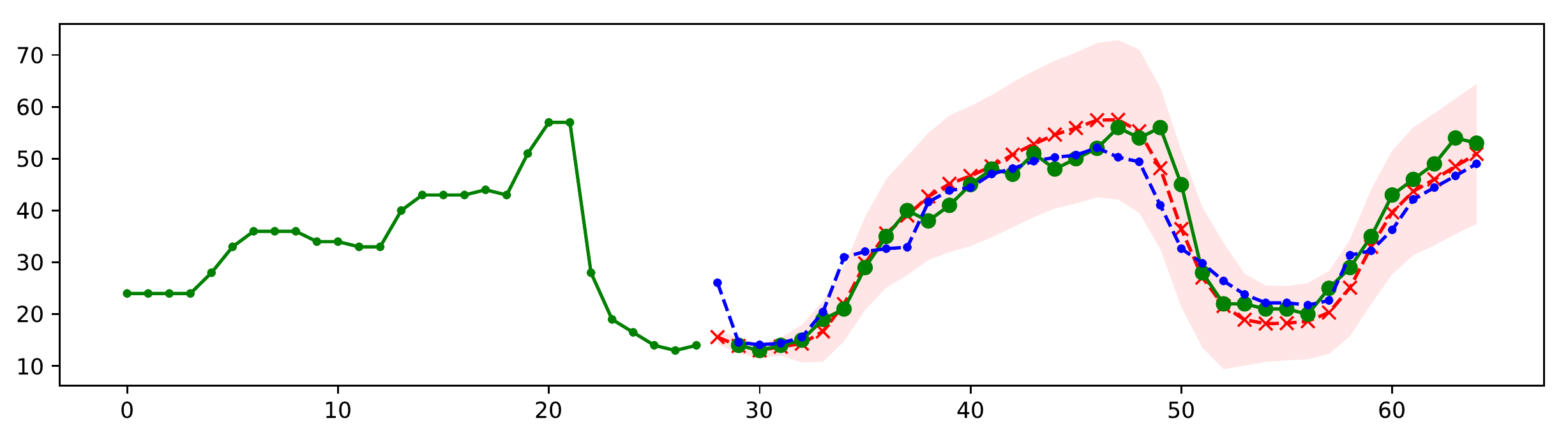}
          \caption{rh2m}
          \label{fig:2}
        \end{subfigure}
        ~
       \begin{subfigure}{0.3\textwidth}
          \includegraphics[width=\textwidth]{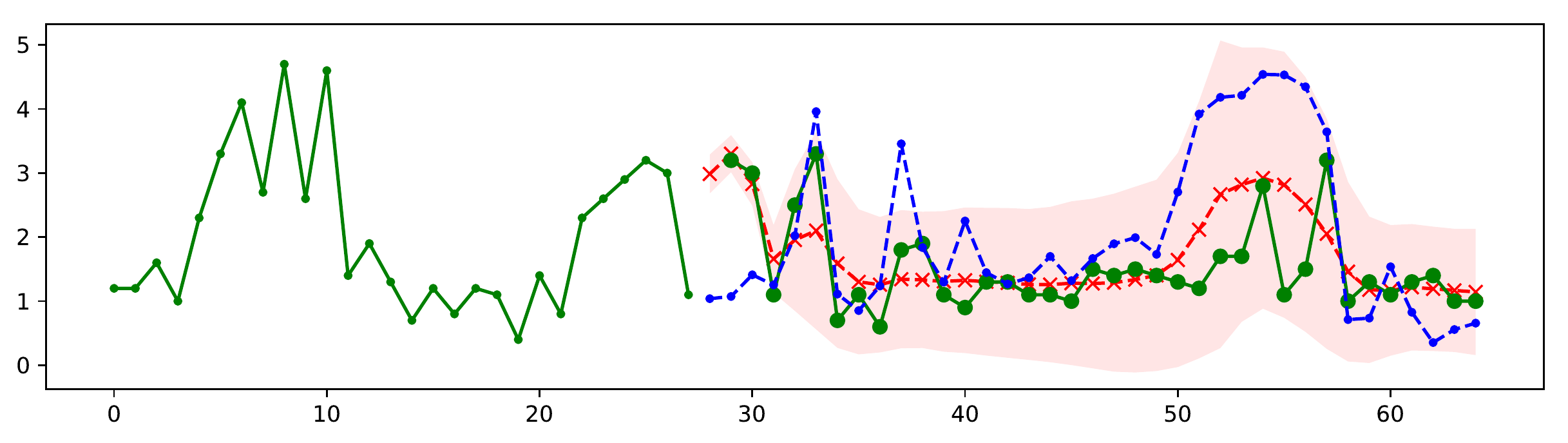}
          \caption{w10m}
          \label{fig:3}
       \end{subfigure}
       \caption{A test sample at one station is chosen to visualize the forecasting of 3 target variables in the future 37 hours. We can see that all predicted points fall into the prediction interval given by $1-z=90\%$.}
       \label{fig:vis}
   \end{figure*}

We take the model DUQ$_{300-300}$ to visualize the quality of the prediction interval. Fig. \ref{fig:vis} illustrates a forecasting instance at one station on a competition day. In each sub-figure, the left green line is the observed meteorological value during the previous 28 hours, the right green line is the ground truth, the blue line is the NWP prediction, the red line is DUQ$_{300-300}$ prediction and the red shaded area is the 90\% prediction interval. A noticeable observation is that the prediction interval does not become wider over time, instead, it presents that the width of the middle part is narrower than both ends particularly for t2m and rh2m.
A reasonable explanation is that meteorological elements largely change during the daytime and become more stable during night time. Having this prediction interval would provide more information for travel/production planning than only point prediction. Another noteworthy point is that because w10m fluctuates sharply, it is more difficult to forecast point estimate precisely, and the prediction interval tends to be wider than t2m and rh2m.

\begin{table*}[!thb]
    \caption{The $SS$ performance of different methods on 9 days. The column P-value compare the best performing method DUQ$_{Esb10}$ with other methods, using one-tail paired T-test.}
    \label{table_SS}
        \begin{center}
        \scalebox{0.8}{\begin{tabular}{cccccccccccc}
        \hline
        Method & $SS_{day1}$ & $SS_{day2}$ & $SS_{day3}$ & $SS_{day4}$ & $SS_{day5}$ & $SS_{day6}$ & $SS_{day7}$ & $SS_{day8}$ & $SS_{day9}$ & $SS_{avg}$& P-value\\
        \hline
        SARIMA & 0.1249 & -1.4632 & -0.2417 & -0.4421 & -0.2631 & -0.2301 & 0.0630 & 0.2015 & -0.4579 & -0.3010& 0.00\\
        SVR & -0.7291 & -0.6342 & -0.1999 & -0.5918 & -1.1230 & -0.8568 & -0.6154 & -0.5123 & -0.5807 & -0.6492& 0.00\\
        GBRT & 0.0221 & 0.1318 & -0.0086 & -0.0396 & -0.0960 & 0.0067 & 0.0772 & 0.0859 & 0.0000 & 0.0199 & 0.00\\
        DUQ$_{50}$ & 0.4813 & 0.4833 & 0.2781 & 0.3053 & 0.4277 & 0.4853 & 0.4609 & 0.4987 & 0.2647 & 0.4095& 0.00\\
        DUQ$_{50-50}$ & 0.4847 & 0.4969 & 0.3088 & 0.4012 & 0.4302 & 0.5051 & \textbf{0.5656} & 0.5502 & 0.3239 & 0.4518& 0.00\\
        DUQ$_{200}$ & 0.5278 & \textbf{0.5088} & 0.2890 & 0.3797 & 0.4479 & 0.5358 & 0.4961 & 0.5235 & 0.3478 & 0.4507& 0.02\\
        DUQ$_{300-300}$ & 0.5220 & 0.5002 & 0.3352 & 0.4067 & 0.4474 & 0.5289 & 0.5324 & 0.5463 & 0.3047 & 0.4582& 0.00\\
        DUQ$_{noNWP}$ & 0.2348 & 0.2992 & 0.0081 & 0.2440 & 0.1630 & 0.3125 & 0.2660 & 0.3003 & -0.1599 & 0.1853 & 0.00\\
        DUQ$_{noOBS}$ & 0.4694 & 0.4744 & 0.2624 & 0.3447 & 0.3925 & 0.4588 & 0.4756 & 0.4901 & 0.3150 & 0.4092 & 0.00 \\
        Seq2Seq$_{MSE}$ & 0.4978 & 0.3934 & 0.2860 & 0.3960 & 0.3965 & 0.4842 & 0.4820 & 0.5138 & 0.3192 & 0.4188& 0.00\\
        Seq2Seq$_{MAE}$ & 0.5314 & 0.4346 & 0.2671 & 0.3980 & 0.4610 & 0.5391 & 0.4711 & 0.5565 & 0.2999 & 0.4399& 0.00\\
        DUQ$_{Esb3}$ (2rd place) & 0.5216 & 0.4951 & 0.3358 & 0.4050 & 0.4627 & 0.5359 & 0.5350 & 0.5664 & \textbf{0.3479} & 0.4673& 0.04\\
        DUQ$_{Esb10}$ & \textbf{0.5339} & 0.4940 & \textbf{0.3516} & 0.4355 & 0.4600 & 0.5575 & 0.5581 & \textbf{0.5776} & 0.3298 & \textbf{0.4776}& -\\
        Model$_{1st}$ & 0.4307 & 0.4847 & 0.3088 & \textbf{0.4572} & \textbf{0.5019} & \textbf{0.5753} & 0.5345 & 0.5726 & 0.3384 & 0.4671& 0.24\\
        \hline
        \end{tabular}}
        \end{center}
    \end{table*}
    
    \begin{table*}[!thb]
    \caption{The $RMSE$ performance of different methods on 9 days. Since $RMSE$ and $SS$ are not fully linear relationship, the counterpart assessment does not reach the optimal at the same time.}
    \label{table_RMSE}
        \begin{center}
        \scalebox{0.75}{\begin{tabular}{cccccccccccc}
        \hline
        Method & $RMSE_{day1}$ & $RMSE_{day2}$ & $RMSE_{day3}$ & $RMSE_{day4}$ & $RMSE_{day5}$ & $RMSE_{day6}$ & $RMSE_{day7}$ & $RMSE_{day8}$ & $RMSE_{day9}$ & $RMSE_{avg}$ & P-value \\
        \hline
        NWP & 7.5923 & 9.7276 & 5.1079 & 5.7335 & 6.1542 & 7.5239 & 6.3647 & 7.0457 & 5.8819 & 6.7924 & 0.00\\
        SARIMA & 7.3017 & 16.5954 & 7.2964 & 8.9968 & 8.0726 & 8.1440 & 7.1722 & 5.8466 & 8.1795 & 8.6228 & 0.00 \\
        SVR & 8.1788 & 10.0436 & 5.9310 & 7.1662 & 7.7576 & 8.4064 & 7.9094 & 8.4749 & 7.5431 & 7.9346 & 0.00 \\
        GBRT & 6.5551 & 8.0321 & 5.6892 & 6.1422 & 6.1083 & 6.5687 & 5.9449 & 6.7122 & 5.9573 & 6.4122 & 0.00 \\
        DUQ$_{50}$ & 3.0432 & 4.5256 & 4.1858 & 4.5445 & 3.0069 & 3.4912 & 3.8087 & 3.2299 & 4.5545 & 3.8211 & 0.00 \\
        DUQ$_{50-50}$ & 2.9013 & 4.2442 & 4.0203 & 3.6641 & 3.2275 & 3.2062 & \textbf{2.6428} & 2.8175 & 4.3089 & 3.4481 & 0.00 \\
        DUQ$_{200}$ & 2.8128 & \textbf{4.0400} & 4.1624 & 4.0732 & 2.8580 & 2.8086 & 3.2870 & 3.1963 & 4.3326 & 3.5079 & 0.03\\
        DUQ$_{300-300}$ & 2.7168 & 4.0615 & 3.8866 & 3.7977 & 2.8083 & 2.9211 & 2.8012 & 2.9784 & \textbf{4.3308} & 3.3669 & 0.16\\
        DUQ$_{noNWP}$ & 5.0371 & 5.5370 & 5.0529 & 4.6819 & 4.1385 & 4.4716 & 5.7058 & 5.8346 & 7.6805 & 5.3489 & 0.00\\
        DUQ$_{noOBS}$ & 3.2170 & 4.8604 & 4.4150 & 4.1303 & 3.5896 & 3.8239 & 3.4992 & 3.2031 & 4.3008 & 3.8933 & 0.00\\
        Seq2Seq$_{MSE}$ & 3.1328 & 5.0769 & 4.1400 & 3.8426 & 3.2040 & 3.3142 & 3.3027 & 3.1785 & 4.6426 & 3.7594 & 0.00 \\
        Seq2Seq$_{MAE}$ & 2.7272 & 5.0933 & 4.2837 & 4.0184 & 2.7888 & 3.0029 & 3.7165 & 2.7935 & 4.6509 & 3.6750 & 0.00 \\
        DUQ$_{Esb3}$ & 2.8000 & 4.4338 & \textbf{3.7054} & 3.7886 & 2.8566 & 2.7890 & 2.7979 & 2.8011 & 4.3310 & 3.3670 & 0.05\\
        DUQ$_{Esb10}$ & \textbf{2.7027} & 4.3341 & 3.7999 & \textbf{3.5743} & \textbf{2.7627} & \textbf{2.6874} & 2.7799 & \textbf{2.7402} & 4.3949 & \textbf{3.3085} & -\\
        \hline
        \end{tabular}}
        \end{center}
    \end{table*}
    
\begin{table*}[!thb]
    \caption {PICP on every day}
    \begin{center}
    \scalebox{0.75}{\begin{tabular}{ccccccccccc}
    \hline
    \makecell{$PICP_{obj}$}& Day 1      & Day 2      & Day 3      & Day 4      & Day 5      & Day 6      & Day 7      & Day 8      & Day 9      & $PICP_{avg}$    \\ \hline
    $PICP_{t2m}$                      &     0.9513 &     0.9351 &     0.8918 &     0.8891 &     0.9594 &     0.9648 &     0.9027 &     0.8945 &     0.9351 &     \textbf{0.9249} \\
    $PICP_{rh2m}$                    &     0.9945 &     0.8702 &     0.7648 &     0.8729 &     0.9621 &     0.9621 &     0.9243 &     0.9243 &     0.9135 &     \textbf{0.9099} \\
    $PICP_{w10m}$                     &     0.9567 &     0.9567 &     0.9081 &     0.9594 &     0.9675 &     0.9621 &     0.9378 &     0.9648 &     0.9540 &     \textbf{0.9519} \\
    \hline
    \end{tabular}}
\end{center}
\label{tab_picp}
\end{table*}

\begin{table}[!thb]
    
    \caption {T-test among single models for DUQ$_{300-300}$}
    \begin{center}
        \scalebox{0.6}{\begin{tabular}{ccc}
        \hline
        ~             & P-value on RMSE$_{avg}$ & P-value on SS$_{avg}$ \\ \hline
        DUQ$_{300-300}$ & -            & -          \\
        DUQ$_{noNWP}$ & 0.00            & 0.00          \\
        DUQ$_{noOBS}$ & 0.00            & 0.00          \\
        Seq2Seq$_{MSE}$  & 0.00         & 0.00       \\
        Seq2Seq$_{MAE}$ & 0.03         & 0.08       \\ \hline
        \end{tabular}}
    \end{center}
    \label{tab_p_value}
\end{table}

\begin{table}[!thb]
\caption {Iterations}
     \scalebox{0.7}{\begin{tabular}{cc}
    \hline
    Methods               & ti (vt$\times$ vi) \\ \hline
    DUQ$_{300-300}$ & 2900 (58*50)                       \\
    Seq2Seq$_{MSE}$                    & 2100 (42*50)                       \\
    DUQ$_{noNWP}$                    & 1950 (39*50)                       \\
    Seq2Seq$_{MAE}$                    & 1850 (37*50)                       \\
    DUQ$_{noOBS}$                    & 1450 (29*50)                       \\
    \hline
    \end{tabular}}
    \label{tab_iter}
\end{table}

\section{CONCLUSIONS AND FUTURE WORKS}
This paper addresses the real-world problem in  weather forecasting which has a profound impact on
our daily life, by introducing a new deep uncertainty quantification (DUQ) method. A novel loss function called negative log-likelihood error (NLE) was designed to train the prediction model, which is capable of simultaneously inferring sequential point estimation and prediction interval.
A noteworthy experimental phenomenon reported in this paper is that training by NLE loss significantly improves the generalization of point estimation. This may provide practitioners with new insights to develop and deploy learning algorithms for related problems such as time series regression. Based on the proposed method and an efficient deep ensemble strategy, state-of-the-art performance on a real-world benchmark dataset of weather forecasting was achieved. The overall method was developed in Keras and was flexible enough to be deployed in the production environment. The data and source codes will be released and can be used as a benchmark for researchers and practitioners to investigate weather forecasting.
Future works will be directed towards architecture improvement (e.g., attention mechanism), automatic hyperparameter-tuning, and theoretical comparison between NLE and MSE/MAE.

\begin{acks}
This work was supported by the Australian Research Council (No. DP150101645), the Natural Science Foundation of China (No. 61773324) and the Fundamental Research Funds for the Central Universities (No. 2682015QM02) .
\end{acks}

\bibliographystyle{ACM-Reference-Format}
\bibliography{sample-base}

\end{document}